\documentclass[twoside]{article}
\usepackage[accepted]{aistatsarxiv}

\newcommand{\ddirichlet}[0]{\mathrm{Dirichlet}}
\newcommand{\dpoisson}[0]{\mathrm{Poisson}}
\newcommand{\dgamma}[0]{\mathrm{Gamma}}

\newcommand{\dbeta}[0]{\mathrm{Beta}}
\newcommand{\dbernoulli}{\mathrm{Bernoulli}}
\newcommand{\duniform}{\mathrm{Uniform}}

\newcommand{\dmultinomial}{\mathrm{Multinomial}}

\newcommand{\cL}{\mathcal{L}}
\newcommand{\E}{\mathbb{E}}
\newcommand{\Eq}{\mathbb{E}_q}

\newcommand{\I}{\mathbb{I}}
\newcommand{\KL}{\mathrm{KL}}

\newcommand{\partiald}[2]{\frac{\partial {#1}}{\partial {#2}}}

\usepackage{times,amsmath,amssymb,natbib}
\usepackage{algorithm,algorithmic}
\usepackage{url,epsfig}

\begin{document}

%

%

\twocolumn[

\aistatstitle{Structured Stochastic Variational Inference}

\aistatsauthor{ Matthew D. Hoffman \And David M. Blei }

\aistatsaddress{ Adobe Research \And Columbia University } ]

\begin{abstract}
Stochastic variational inference makes it possible to approximate
posterior distributions induced by large datasets quickly using
stochastic optimization.  The algorithm relies on the use of
fully factorized variational distributions. However, this
``mean-field'' independence approximation limits the fidelity of the
posterior approximation, and introduces local optima. We show how to
relax the mean-field approximation to allow arbitrary dependencies
between global parameters and local hidden variables, producing better
parameter estimates by reducing bias,
sensitivity to local optima, and sensitivity to hyperparameters.
\end{abstract}

\section{Introduction}
Hierarchical Bayesian modeling is a powerful framework for learning
from rich data sources. Unfortunately, the intractability of
the posteriors of rich models drives practitioners to resort to
approximate inference algorithms such as mean-field variational
inference or Markov chain Monte Carlo (MCMC). These classes of methods
have complementary strengths and weaknesses---MCMC methods have strong
asymptotic guarantees of unbiasedness but are often slow, while
mean-field variational inference is often faster but tends to
misrepresent important qualities of the posterior of interest and is
more vulnerable to local optima.  Incremental versions of both methods
based on stochastic optimization have been developed that are
applicable to large datasets \citep{Welling:2011,Hoffman:2013}.

In this paper we focus on variational inference. In particular, we are
interested in using variational inference to estimate the parameters
of high-dimensional Bayesian models with highly multimodal posteriors,
such as mixture models, topic models, and factor models. We will be
less concerned with uncertainty estimates, since they are difficult to
trust and interpret in this setting.

Mean-field variational inference approximates the intractable
posterior distribution implied by the model and data with a factorized
approximating distribution in which all parameters are
independent. This mean-field distribution is then tuned to minimize
its Kullback-Leibler divergence to the posterior, which is equivalent
to maximizing a lower bound on the marginal probability of the
data. The restriction to factorized distributions makes the
problem tractable, but reduces the fidelity of the
approximation and introduces local optima \citep{Wainwright:2008}.

A partial remedy is to weaken the mean-field factorization by
restoring some dependencies, resulting in ``structured'' mean-field
approximations \citep{Saul:1996a}. The applicability, speed,
effectiveness, and ease-of-implementation of standard structured
mean-field algorithms is limited because the lower bound implied by
the structured distribution must be available in closed form.

More recent work manages these intractable variational
lower bounds using stochastic optimization, which allows one to optimize
functions that can only be computed approximately. For example,
\citet{Ji:2010} use mean-field approximations to the posteriors of
``collapsed'' models where some parameters have been analytically
marginalized out, \citet{Salimans:2013} apply a structured
approximation to the posterior of a stochastic volatility model, and
\citet{Mimno:2012} use a structured approximation to the posterior of
a collapsed model. 

In parallel, \citet{Hoffman:2013} proposed the stochastic variational
inference (SVI) framework, which uses stochastic optimization to apply
mean-field variational inference to massive datasets.  SVI splits
the unobserved variables in a hierarchical model into global
parameters $\beta$ (which are shared across all observations) and
groups of local hidden variables $z_1,\ldots,z_N$ (each of which is
specific to a small group of observations $y_n$). The goal is to
minimize the Kullback-Leibler (KL) divergence between a tractable
approximating distribution $q(z,\beta)$ over the local and global
parameters and the true posterior $p(z,\beta|y)$ over those
parameters.  SVI approximates posteriors much more quickly than
traditional batch variational inference algorithms.

Like batch variational inference, SVI relies on the mean-field
approximation, which requires that $q$ factorize as $(\prod_k
q(\beta_k))\prod_{n,m} q(z_{n,m})$; that is, SVI approximates the
joint posterior $p(z,\beta|y)$ with a distribution $q$ that cannot
represent any dependencies between random variables.
\citet{Mimno:2012} proposed a variant for latent Dirichlet allocation
that restores dependencies between sets of local hidden variables
$z_{n,1:M}$ so that the approximating distribution has the form
$q(z,\beta)=(\prod_k q(\beta_k))\prod_n q(z_{n,1:M})$, improving $q$'s
ability to approximate the posterior $p(z,\beta|y)$. But their method
still breaks the dependence between the global and local variables
$\beta$ and $z$.

In this paper, we introduce structured stochastic variational
inference (SSVI), a generalization of the SVI framework that can
restore the dependence between global and local variables. SSVI
approximates the posteriors $p(z, \beta|y)$ of a wide class of models
with distributions of the form $q(z,\beta)=(\prod_kq(\beta_k))\prod_n
q(z_n|\beta)$, allowing for arbitrary dependencies between $\beta$ and
$z_n$.

In experiments on three models and datasets, we find that in all cases
restoring these dependencies allows SSVI to find qualitatively and
quantitatively better parameter estimates, avoiding the local optima
and sensitivity to hyperparameters that plague mean-field variational
inference.

\begin{figure}[t]
  \centerline{\includegraphics[width=0.5\textwidth]{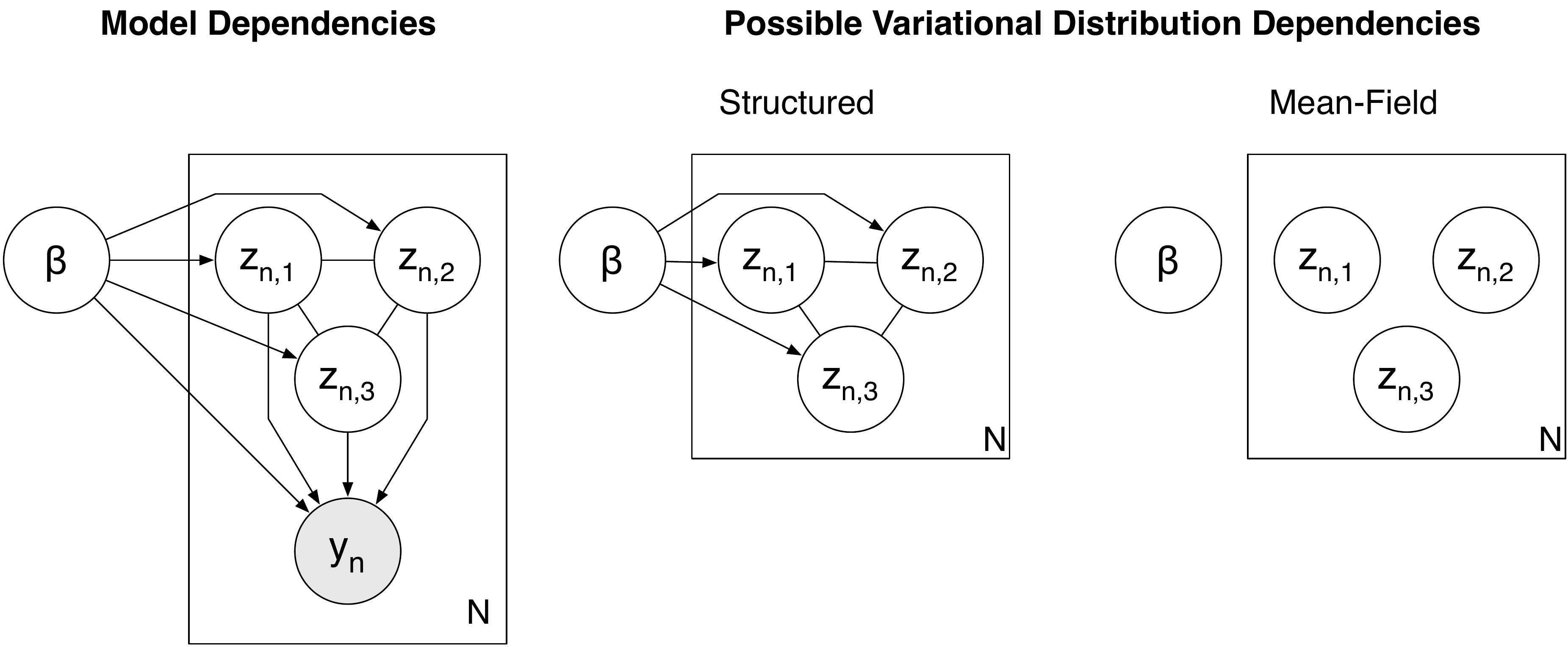}}
  \caption{Left: Graphical model showing the independence structure of
    models in our framework. We make no assumptions on dependencies
    between the elements of $z_n$, so the elements of $z_n$ are all
    connected. Here we show only three local $z$ variables, but there
    could be more or fewer.  Middle: Graphical model of structured
    variational distributions $q(\beta, z)$ that are possible in our
    framework. Right: Graphical model of mean-field variational
    distributions.}
\label{fig:modelgm}
\end{figure}

\section{Structured Stochastic Variational Inference}
\label{sec:ssvi}
In this section, we will present two SSVI algorithms. We first review
the class of models to which SSVI can be applied and the variational
distributions that it employs. 

\subsection{Model Assumptions}
\label{sec:modelassumptions}
As in SVI \citep{Hoffman:2013}, we assume we have $N$ groups of
observations $y_{1:N}$ and a probability model that factorizes as
$\textstyle p(y, z, \beta) = p(\beta) \prod_n p(y_n, z_n | \beta).$
The independence structure of such a model is visualized in figure
\ref{fig:modelgm}.  The global parameters $\beta$ are shared across
all observations, and the local hidden variables $z_{1:N}$ are
conditionally independent of one another given the global parameters
$\beta$.

We will restrict our attention to conditionally conjugate models. We
assume that the prior $p(\beta)$ is in a tractable exponential
family $p(\beta) = h(\beta)\exp\{\eta\cdot t(\beta) - A(\eta)\}$.
The base measure $h$ and log-normalizer $A$ are scalar-valued
functions, $\eta$ is a vector of natural parameters, and $t(\beta)$ is
a vector-valued sufficient statistic function. We futher assume that
the joint likelihood of the local variables $y_n$ and $z_n$ given
$\beta$ is of the form $p(y_n, z_n | \beta) = \exp\{t(\beta)\cdot
\eta_n(y_n, z_n) + g_n(y_n, z_n)\}$ (where $g_n$ is a real-valued
function and $\eta_n$ is a vector-valued function).

This form for $p(y_n, z_n, \beta)$ includes all conjugate pairs of
distributions $p(\beta)$, $p(y_n, z_n|\beta)$ \citep{Gelman:2013};
that is, it is the most general family of distributions for which
the conditional $p(\beta|y,z)$ is in the same family as the prior
$p(\beta)$. This conditional is
\begin{equation}
\begin{split}
\label{eq:model}
p(\beta | y, z) =
h(\beta)\exp\{(&\textstyle\eta + \sum_n \eta_n(y_n, z_n))\cdot t(\beta) - 
\\ &\textstyle A(\eta + \sum_n \eta_n(y_n, z_n))\}.
\end{split}
\end{equation}
%
These restrictions are a weaker version of those imposed by
\citet{Hoffman:2013}; the difference is that we make no
assumptions about the tractability of the conditional distributions
$p(z_n | y_n, \beta)$ or $p(z_{n,m}|y_n,z_{n,\backslash
  m},\beta)$. This work is therefore applicable to any model that fits
in the SVI framework, including mixture models, LDA, hidden Markov
models (HMMs), factorial HMMs, Kalman filters, factor analyzers,
probabilistic matrix factorizations, hierarchical linear regression,
hierarchical probit regression, and many other hierarchical
models. Unlike SVI, it can also address models without tractable local
conditionals, such as multilevel logistic regressions
\citep{Gelman:2007} or the correlated topic model \citep{Blei:2006b}.

\subsection{Approximating Distribution}
Our goal is to approximate the intractable posterior $p(z, \beta
| y)$ with a distribution $q(z,\beta)$ in some restricted, tractable
family. We will choose a $q$ distribution from this family by solving
an optimization problem, minimizing the Kullback-Leibler (KL)
divergence between $q(z,\beta)$ and the posterior
$p(z,\beta | y)$.

The simplest approach is to make the mean-field approximation,
restricting $q$ to factorize so that $q(z,\beta) =
q(\beta)\prod_n\prod_m q(z_{n,m}).$ This restriction dramatically
simplifies the form of the KL-divergence between $q$ and the
posterior, but this simplicity comes at a price. Every dependence that
we break to make $q$ easier to work with makes $q$ less able to
closely approximate the posterior $p(z, \beta|y)$. Breaking
dependencies may also introduce additional local minima into the KL
divergence between $q$ and the posterior; imposing independence
assumptions places nonconvex constraints on the dual of the solution
space, which may block the path from a bad solution to a good one
\citep{Wainwright:2008}.

Structured mean-field partially relaxes the mean-field
independence restriction \citep{Saul:1996a}. Traditional structured
mean-field algorithms require the practitioner to identify and exploit
some model-specific structure; for example, \citet{Ghahramani:1997}
exploited the availability of dynamic programming algorithms for HMMs
to derive a structured mean-field algorithm for factorial HMMs.

\citet{Mimno:2012} proposed a structured stochastic variational
inference algorithm for latent Dirichlet allocation that depends less
on model-specific structure, placing no restrictions on the joint
distributions $q(z_{n,1:M})$ so that $q(z, \beta)$ factorizes as $q(z,
\beta) = (\prod_k q(\beta_k))\prod_n q(z_n).$ The optimal $q(z_n)$ may
not be tractable to normalize, but it can still be sampled from using
Markov chain Monte Carlo (MCMC), which is all that is necessary to
generate a stochastic natural gradient for a stochastic variational
inference algorithm. The result was a significant improvement in the
quality of the inference algorithm's ability to obtain high-quality
estimates of model parameters. However, the approximate posterior of
\citet{Mimno:2012} still breaks the dependence between global
parameters $\beta$ and local hidden variables $z$.

We introduce a framework for \emph{structured stochastic variational
  inference} (SSVI) algorithms that restore the dependence between
$\beta$ and $z$. Our variational distribution $q$ is of the form
\begin{equation}
\begin{split}
\textstyle
q(z, \beta) = (\prod_k q(\beta_k))\prod_n q(z_n | \beta).
\end{split}
\end{equation}
The only remaining factorization we impose is between the elements of
$\beta$; the conditional independence between the $z_n$s
given $\beta$ is implied by the model in equation \ref{eq:model}.

We will restrict $q(\beta)$ to be in the same exponential family as
the prior $p(\beta)$, so that $q(\beta) = h(\beta)\exp\{\lambda\cdot
t(\beta) - A(\lambda)\}.$ $\lambda$ is a vector of free parameters
that controls $q(\beta)$.
We also require that any dependence under $q$ between $z_n$ and $\beta$ be
mediated by some vector-valued function $\gamma_n(\beta)$, so that we
may write $q(z_n|\beta)=q(z_n|\gamma_n(\beta))$. 

This form for $q$ allows for rich dependencies between nearly all
variables in the model. This comes at a cost, however. In mean-field
variational inference, we proceed by maximizing a lower bound on the
marginal probability of the data; this is equivalent to minimizing the
KL divergence from $q$ to the posterior \citep{Bishop:2006}. However,
this lower bound contains expectations that become impossible to
compute when we allow $z_n$ to depend on $\beta$ in $q$. This issue
may seem insurmountable, but even though we cannot compute the
variational lower bound, we can still optimize it using stochastic
optimization.

\subsection{The Structured Variational Objective \label{sec:requirements}}
Our goal is to find a distribution $q(\beta, z)$ that has low KL
divergence to the posterior $p(\beta, z|y)$.  The KL divergence
between $q$ and the full posterior is
\begin{equation}
\begin{split}
\KL(q_{ z,\beta}||p_{ z, \beta| y}) 
&= -\Eq[\log p(y,  z, \beta)] + 
\\ &\qquad\Eq[\log q( z,  \beta)]
 + \log p(y).
\end{split}
\end{equation}
Because the KL divergence must be non-negative, this yields the evidence
lower bound (ELBO)
\begin{align}
\label{eq:elbo}
\textstyle
\cL &\equiv \Eq[\log p(y, z, \beta)]
- \Eq[\log q( z,  \beta)]
\nonumber
\\ &\textstyle = \Eq[\log \frac{p(\beta)}{q(\beta)}]
+ \sum_n \Eq[\log \frac{p(y_n, z_n | \beta)}{q(z_n | \beta)}]
\nonumber
\\ &\textstyle = \int_\beta q(\beta)\big(\log \frac{p(\beta)}{q(\beta)} +
\\ & \textstyle\quad + \sum_n\int_{z_n}q(z_n|\beta)
\log \frac{p(y_n, z_n|\beta)}{q(z_n|\beta)}dz_n\big)d\beta
\le \log p(y),
\nonumber
\end{align}
We used the conditional independence structure assumed
in section \ref{sec:modelassumptions} to break $\log p(y, z|\beta)$ into a
sum over $n$. Our goal is to maximize the ELBO subject to some
restrictions on $q$.



Before describing the form we choose for $\gamma_n(\beta)$, we first
note that the second integral in equation \ref{eq:elbo} is itself a
lower bound on the marginal probability of the $n$th group of observations:
\begin{align}
\label{eq:localbound}
&\textstyle\int_{z_n} q(z_n|\beta)\log\frac{p(y_n, z_n|\beta)}{q(z_n|\beta)}dz_n
\\ &= -\KL(q_{ z_n|\beta}||p_{ z_n|y_n,\beta}) + \log p(y_n|\beta)
\le \log p(y_n|\beta).
\nonumber
\end{align}
Thus, for any particular value of $\beta$ we can maximize the global
ELBO over $q(z_n|\beta)$ by minimizing the KL divergence between
$q(z_n|\beta)$ and $p(z_n|y_n,\beta)$. We will assume that the
function $\gamma_n(\beta)$ that controls
$q(z_n|\beta)=q(z_n|\gamma_n(\beta))$ is defined to do just that, so
that $\gamma_n(\beta)$ is at a local maximum of this ``local ELBO'',
i.e.,
\begin{equation}
\begin{split}
\label{eq:gammastar}
\textstyle
\nabla_{\gamma_n} \int_{z_n}q(z_n|\gamma_n(\beta))
\log \frac{p(y_n, z_n|\beta)}{q(z_n|\gamma_n(\beta))}dz_n
= 0.
\end{split}
\end{equation}
The function $\gamma_n(\beta)$ may be implicit; e.g., it might be
evaluated by solving an optimization problem.

\subsection{Algorithms}

Our algorithm for optimizing the ELBO from equation \ref{eq:elbo} is
summarized in algorithm \ref{alg:structured}. Each iteration, we
sample the global parameters $\beta$ from $q(\beta)$. We then compute
the parameters $\gamma_n(\beta)$ to each local variational
distribution $q(z_n|\gamma_n(\beta))$ that maximize the local ELBO
$\Eq[\log p(y_n,z_n|\beta) - \log q(z_n|\beta)|\beta]$ and compute an
unbiased estimate $\hat\eta_n$ of $\Eq[\eta_n(y_n, z)|\beta]$, the expected
value of $\eta_n(y_n, z)$ with respect to the maximized local
ELBO. Finally, we update $\lambda$ with some step size $\rho$ so that:
\begin{equation}
\begin{split}
\label{eq:ssmfupdate}
\textstyle
\lambda' = (1-\rho)\lambda + \rho(\eta + 
V(\beta, \lambda)\sum_n\hat\eta_n).
\end{split}
\end{equation}
where $V(\beta, \lambda)$ is a matrix that is defined in terms of the
cumulative distribution functions (CDFs) quantile functions
(inverse-CDFs) of $q(\beta)$. Defining the CDF
$Q_k(\beta_k)\equiv\int_{-\infty}^{\beta_k} q(\beta_k')d\beta_k'$ and
the quantile function $R_k(Q_k(\beta_k))\equiv\beta_k$, $V(\beta,
\lambda)$ is defined as the product of two matrices: the inverse of
the second derivative of the log-normalizer $A$ of $q$, and the
Jacobian of $t(R(Q(\beta)))$ with respect to $\lambda$:
\begin{equation}
\begin{split}
\label{eq:vdef}
V(\beta, \lambda)\equiv \nabla^2_\lambda A(\lambda)^{-1}
\nabla_\lambda R(Q(\beta)) \nabla_\beta t(\beta)^\top.
\end{split}
\end{equation}
Equation \ref{eq:ssmfupdate} is derived in appendix
\ref{sec:derivation} as a stochastic natural gradient update of the
SSVI ELBO. It resembles the standard SVI update, with two differences:
we use a sample from $q(\beta)$ instead of $\Eq[t(\beta)]$ when
estimating $\hat\eta_n$, and we multiply $\hat\eta_n$ by the matrix
$V(\beta,\lambda)$.

The matrix $V$ appears because of the dependence between $\beta$ and
$\hat\eta$. $V$'s expected value is the identity matrix, since
$\Eq[\nabla_\lambda t(R(Q(\beta)))] =\nabla_\lambda\Eq[t(\beta)]
=\nabla^2_\lambda A(\lambda).$ (The last identity follows from
$q(\beta)$ being in an exponential family.) So we can
decompose $V\hat\eta$ as
\begin{equation}
\begin{split}
V\hat\eta = (V - \Eq[V])(\hat\eta - \Eq[\hat\eta]) + \hat\eta.
\end{split}
\end{equation}
If the covariance-like expression $(V - \Eq[V])(\hat\eta - \Eq[\hat\eta])$ is
small (either because the variance of $V$ or $\hat\eta$ is small or
because $V$ and $\hat\eta$ do not depend strongly on one another), we can
neglect it without introducing much bias. This suggests the simpler
update (labeled SSVI-A in algorithm \ref{alg:structured})
\begin{equation}
\begin{split}
\label{eq:ssmfaupdate}
\textstyle
\lambda' = (1-\rho)\lambda + \rho(\eta + \sum_n\hat\eta_n),
\end{split}
\end{equation}
which avoids the difficulty and (modest) expense of
computing derivatives of quantiles and log-normalizers.
 


 
\begin{algorithm}[t]
\begin{algorithmic}[1]
  \STATE Initialize $t=1$, initialize $\lambda^{(0)}$ randomly.\\
  \REPEAT
  \STATE Compute step size $\rho^{(t)}=st^{-\kappa}$.
  \STATE Set $N_t=\min \{t, N\}$.
  \STATE Sample global parameters $\beta^{(t)}$ from $q(\beta)$.
  \STATE Compute the local variational parameters $\gamma_n(\beta^{(t)})$.
  \STATE Compute an estimate $\hat\eta_n$ such that \\$\E[\hat\eta_n]=\int_{z_n} q(z_n|\gamma_n(\beta^{(t)}))\eta_n(y_n,z_n)dz_n$.
  \STATE {\bf Option 1: SSVI.} \\
Set $\lambda^{(t)}=$\\$\ (1-\rho^{(t)})\lambda^{(t-1)} + \rho^{(t)}
  (\eta + V(\beta^{(t)}, \lambda^{(t)})\sum_n \hat\eta_n)$. (See equation
\ref{eq:vdef} for definition of $V(\beta, \lambda)$.)
  \STATE {\bf Option 2: SSVI-A.} \\
Set $\lambda^{(t)}=$\\$\ (1-\rho^{(t)})\lambda^{(t-1)} + \rho^{(t)}
  (\eta + \sum_n \hat\eta_n).$
  \UNTIL{convergence}
\end{algorithmic}
\caption{Structured stochastic variational inference (SSVI). \label{alg:structured}}
\end{algorithm}
\subsection{A matrix of approaches}
We are free to choose any form for $q(z_n|\beta)$ and any unbiased
estimator $\hat\eta$ that we want; different choices have different
properties:
\paragraph{Exact conditional:} $q(z_n|\beta) = p(z_n|y_n,\beta)$. Setting
  $q(z_n|\beta)$ to the local conditional distribution
  results in the best possible local ELBO. But this conditional may be
  intractable, in which case one must fall back on MCMC to estimate
  $\Eq[\eta_n(y_n,z_n)|\beta]$. With this choice SSVI directly
  minimizes the KL divergence between $q(\beta)$ and the
  \emph{marginal} posterior $p(\beta|y)$, integrating out the
  ``nuisance parameters'' $z$.
\paragraph{Mean-field:} $q(z_n|\beta)=\prod_m q(z_{n,m}|\beta)$. This choice
  may be computationally cheaper, and it will often make it possible to
  compute $\Eq[\eta_n(y_n,z_n|\beta)]$ in closed form, but it reduces
  the algorithm's ability to approximate the marginal posterior
  $p(\beta|y)$. That said, this is still better than a full mean-field
  approach where one also breaks the dependencies between $z$ and
  $\beta$.
\paragraph{Non-bound-preserving approaches:} Although we derived SSVI
  assuming that $q(z_n|\beta)$ is chosen to maximize the
  local ELBO $\cL_{n}$, one could obtain a distribution over
  $z_n$ in other ways. For example, for latent Dirichlet
  allocation one could adapt the CVB0 method of \citet{Asuncion:2009}
  to use a fixed value of $\beta$, resulting in an algorithm akin to
  that of \citet{Foulds:2013}.

All of these choices of local variational distribution could also be
used in a traditional mean-field setup where
$q(\beta,z)=q(\beta)q(z)$, or as part of a variational maximum a
posteriori (MAP) estimation algorithm. So for any model that enjoys
conditional conjugacy we have a matrix of possible variational
inference algorithms: we can match any ``E-step'' (e.g. mean-field or
sampling from the exact conditional) used to approximate
$p(z_n|y_n,\beta)$ with any ``M-step'' (e.g. MAP, mean-field, SSVI,
SSVI-A) used to update our approximation to $p(\beta|y)$.  

\subsection{Extensions}
There are several ways in which the basic algorithms presented above
can be extended:
\paragraph{Subsampling the data.}
As in SVI, we can compute an unbiased estimate of the sum over $n$ in
equation \ref{eq:ssmfupdate} by only computing $\hat\eta_n$ for some
randomly sampled subset of $S$ observations, resulting in the update
\begin{equation}
\begin{split}
\textstyle
\lambda' = (1-\rho)\lambda + \rho(\eta + 
V(\beta, \lambda)\frac{N}{S}\sum_n\hat\eta_n).
\end{split}
\end{equation}
For large datasets, the reduced computational effort of only looking
at a fraction of the data far outweighs the noise that this
subsampling introduces.  Taking a cue from the recent work of
\citet{Broderick:2013b} and \citet{Wang:2012}, we suggest gradually
ramping up the multiplier $N$ over the course of the first sweep over
the dataset.

\paragraph{Hyperparameter updates and parameter hierarchies.}
As in the mean-field stochastic variational inference framework of
\citet{Hoffman:2013}, we can optimize any hyperparameters in our model
by taking 
steps in the direction of the
gradient of the ELBO with respect to those hyperparameters. We can
also extend the framework developed in this paper to models with
hierarchies of global parameters as in appendix A of
\citep{Hoffman:2013}.

\section{Related Work}
The idea of sampling from global variational distributions to optimize
intractable variational inference problems has been proposed
previously in several contexts. \citet{Ji:2010, Nott:2012,
  Gerrish:2013, Paisley:2012a}, and \citet{Ranganath:2014} proposed
sampling without a change of variables as a way of coping with
non-conjugacy. \citet{Kingma:2014} and \citet{Titsias:2014} proposed
methods that do use a change of variables, although their methods
focus more on speed and/or dealing with nonconjugacy than on improving
the accuracy of the variational approximation. \citet{Salimans:2013}
also suggest using a change of variables as a way of dramatically
reducing the variance of a stochastic gradient estimator.

Although some of the above methods use stochastic optimization to
improve the quality of the mean-field approximation, there are major
differences between these methods and SSVI. \citet{Ji:2010} apply
their method to models where some parameters have been analytically
marginalized out, but do not consider explicitly structured
variational distributions. Also, in our informal experiments we found
that the variance of their gradient estimator was unacceptably high
for high-dimensional problems.

\citet{Salimans:2013} apply their stochastic linear regression method
to structured variational distributions in which the natural
parameters of lower-level variational distributions are explicit
functions of draws from higher-level variational distributions. In
comparison, the implicit form we choose for the conditional
variational distribution $q(z|\beta)$ allows for more complicated
dependencies between global and local parameters. Also, their
regression-based approach requires storing and multiplying matrices
that become impractically large for high-dimensional problems.

A final difference is that the papers mentioned above do not exploit
conjugacy relationships, which are central to the ease of
implementation and efficiency of SSVI-A.

Two other related algorithms are due to \citet{Mimno:2012} (which SSVI
generalizes) and \citet{Wang:2012}. The algorithm of \citet{Wang:2012}
also uses sampling and conjugacy relationships to attempt to restore
the dependencies broken in mean-field algorithms, although their method
lacks guarantees of convergence or correctness.

\section{Experiments}

In this section we empirically evaluate SSVI and SSVI-A's ability to
estimate parameters for three hierarchical Bayesian models. In each
case, we find that relaxing the mean-field approximation allows SSVI
and SSVI-A to find significantly better parameter estimates than
mean-field.  We also find evidence suggesting that this superior
performance is primarily due to SSVI/SSVI-A's ability to avoid local
optima.

\begin{figure*}[t]
  \centerline{\hspace{0.1in}\includegraphics[width=0.9\textwidth]{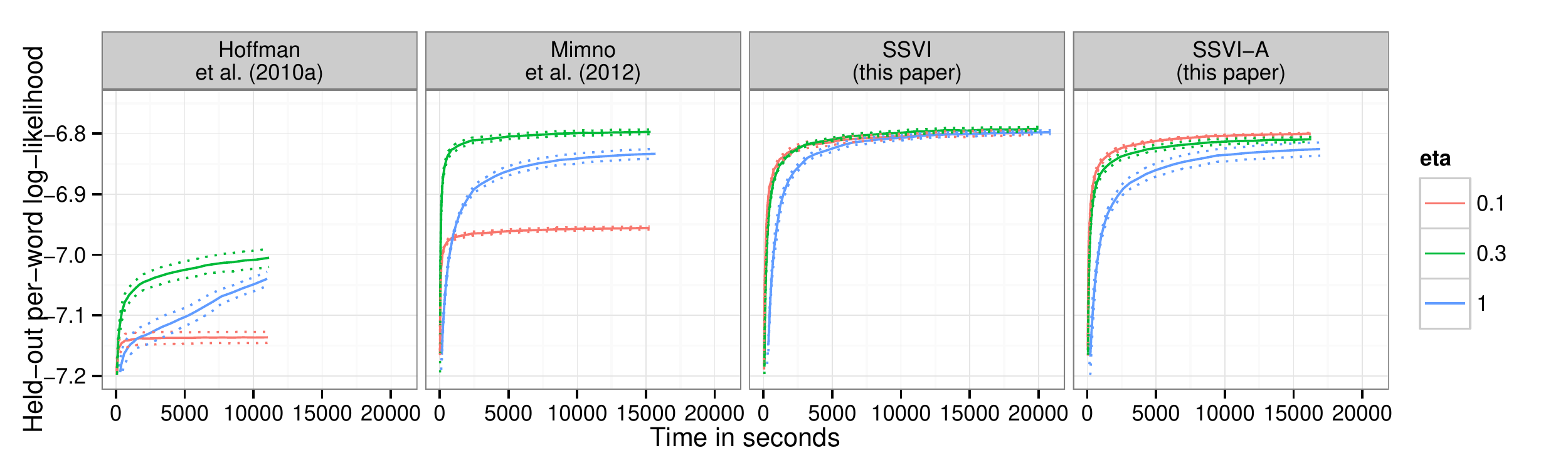}}
  \caption{Predictive accuracy for various algorithms and
    hyperparameter settings as a function of wallclock time when
    fitting LDA to 3.8 million Wikipedia articles. Solid lines show
    average performance across five runs, dotted lines are drawn one
    standard deviation above and below the mean. Each algorithm ran
    for two sweeps over the corpus.
  }
\label{fig:ldaplot}
\end{figure*}

\subsection{Latent Dirichlet allocation}
We evaluated the quality of parameter estimates from SSVI and SSVI-A
on the latent Dirichlet allocation (LDA) topic model fit to the
3,800,000-document Wikipedia dataset from \citep{Hoffman:2013}. We
compared with full mean-field stochastic variational inference \citep{Blei:2003b,
  Hoffman:2010a}, a mean-field M-step with a Gibbs sampling E-step
\citep{Mimno:2012}, SSVI with Gibbs, and SSVI-A with Gibbs.  Results
for other E-step/M-step combinations are in appendix \ref{sec:ldamatrix}.

To speed up learning, each update we subsample a minibatch of 1,000
documents rather than analyzing the whole dataset each iteration.  We
also experimented with various settings of the hyperparameters
$\alpha$ and $\eta$, which mean-field variational inference for LDA is
known to be quite sensitive to \citep{Asuncion:2009}. For all
algorithms we used a step size schedule $\rho^{(t)}=t^{-0.75}$.


We held out a test set of 10,000 documents, and periodically evaluated
the average per-word marginal log probability assigned by the model to
each test document, using the expected value under the variational
distribution as a point estimate of the topics. We estimated marginal
log probabilities with a Chib-style estimator \citep{Wallach:2009a}.

Figure \ref{fig:ldaplot} summarizes the results for $\alpha=0.1$,
which yielded the best results for all algorithms. The method of
\citet{Mimno:2012} outperforms the online LDA algorithm of
\citet{Hoffman:2010a}, but both methods are very sensitive to
hyperparameter selection. SSVI achieves good results regardless of
hyperparameter choice. SSVI-A's performance is very slightly worse
than that of SSVI.


We also evaluated the stochastic gradient Riemannian Langevin dynamics
(SGRLD) algorithm for LDA, which \citet{Patterson:2013} found
outperformed the Gibbs-within-SVI method of \citet{Mimno:2012}. We
experimented with various hyperparameter settings, including the
optimal values reported by \citet{Patterson:2013}.  The best per-word
marginal log probability achieved by SGRLD was $-6.83$. SGRLD thus
outperforms standard SVI regardless of hyperparameters, but only
outperforms Gibbs-within-SVI for some hyperparameter settings, and
achieves performance comparable to that of SSVI and SSVI-A.
This suggests that the poor
performance of Gibbs-within-SVI in the experiments of
\citet{Patterson:2013} may be due to their setting 
$\eta=0.1$.

\paragraph{Computational costs:}

With mini-batches of 1000 documents the computational costs of SSVI
and SSVI-A were comparable to those of Gibbs-within-SVI: SSVI took
about 30\% longer than SSVI to analyze the same number of
documents, and SSVI-A's speed is very close to SSVI. Using Gibbs
sampling in the ``E-step'' rather than variational inference costs
about 50\% more than a mean-field E-step.  We found that SGLRD was
significantly slower per document than either SSVI or SSVI-A, although
this may be due to implementation differences; in principle the
methods should have comparable costs.

\paragraph{Local optima:}
The improved performance of SSVI and SSVI-A over SVI might be because
SSVI/SSVI-A optimize an objective function that more closely
approximates the KL divergence between $q(\beta)$ and $p(\beta|y)$
than the MF objective does.  But it could also be because the MF
objective includes undesirable local optima that the SSVI
objective does not, and so SVI cannot help getting stuck in these
local optima. Our experiments above show that SSVI and SSVI-A
consistently find better parameter estimates than MF even with
multiple restarts, but this does not rule out local optima as an
explanation---even with many restarts it might be very difficult for
MF to find a good local optimum.

\begin{figure}[t]
  \centerline{\hspace{0.1in}\includegraphics[width=0.5\textwidth]{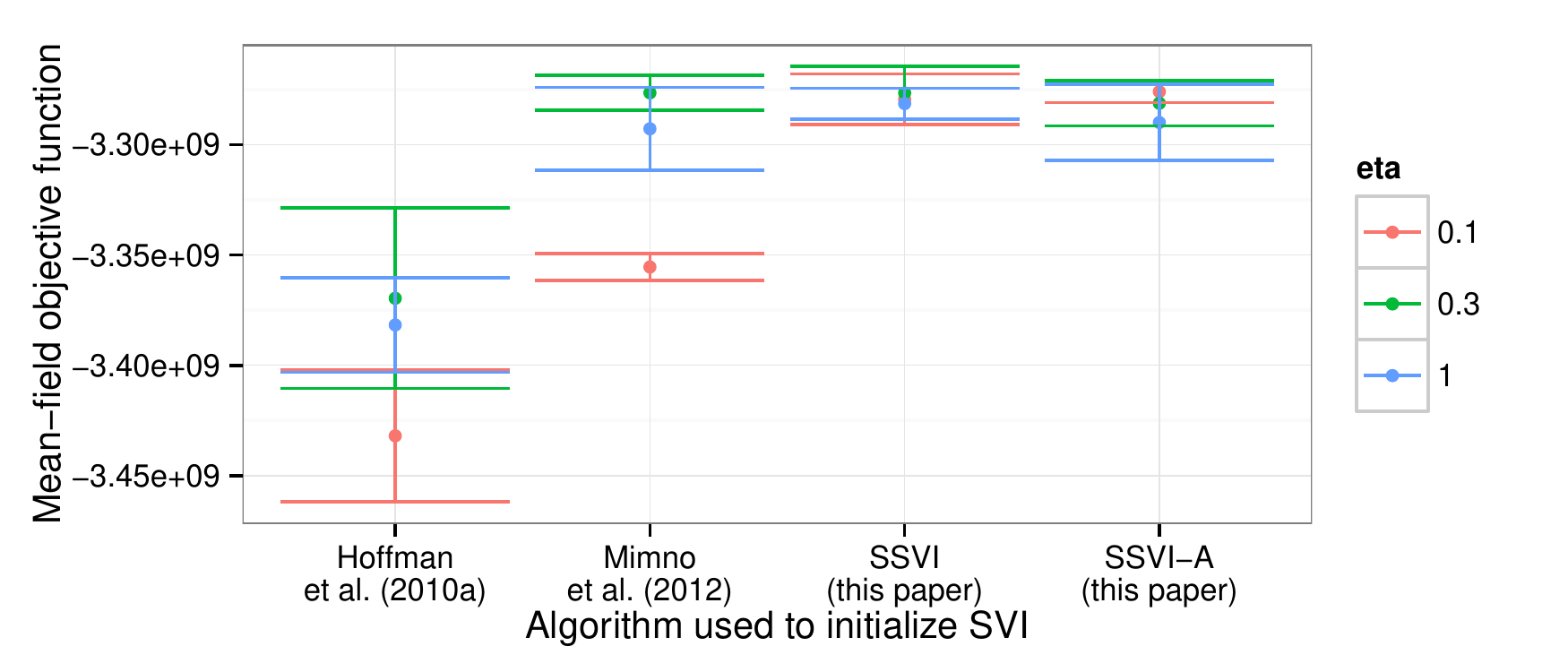}}
  \caption{ELBOs obtained by initializing SVI with the variational
    parameters obtained by four algorithms. Points denote ELBOs
    averaged over five runs, error bars denote the range of ELBOs
    obtained.}
\label{fig:bounds}
\end{figure}

To test this question, we intialized mean-field SVI with the
variational parameters found by SVI, Gibbs-within-SVI, SSVI, and
SSVI-A, and did another sweep through the Wikipedia dataset. Figure
\ref{fig:bounds} plots the mean-field ELBOs obtained by running SVI
from each initialization. SVI initialized with the result from the
structured algorithms finds a much better local optimum of the ELBO on
the training set than SVI initialized randomly.

This result suggests that the main weakness of mean-field methods may
not be the inability of factorized distributions to adequately
approximate the posterior, but the difficulty of finding a good local
optimum of the ELBO (at least in high-dimensional, multimodal
problems). Conversely, the improved performance of the structured
methods may be due to the relative lack of nasty local optima in the
structured ELBOs. That the structured methods find variational
distributions that are also good in the mean-field setting suggests
that the structured ELBO may resemble a smoother version of the
mean-field ELBO.

\subsection{Dirichlet process mixture of Bernoullis}
\label{sec:dpmob}
We used a synthetic data experiment to compare mean-field and
SSVI-A's ability to correctly approximate the posterior of (a finite
approximation to) a Dirichlet process mixture of Bernoullis model. The
data were generated according to the process
$\nonumber
\textstyle \pi \sim \ddirichlet(\frac{\alpha}{K});
\quad \phi_{k,d} \sim \dbeta(1, 1);
\quad z_n \sim \dmultinomial(\pi);
\quad y_{n,d} \sim \dbernoulli(\phi_{z_n,d})$,
where the size of $\pi$ is $K=100$ and the hyperparameter
$\alpha=20$. Each observation $y_n$ is a 100-dimensional binary
vector, and 1000 such vectors were sampled.  The model ultimately used
only 56 of the 100 mixture components to generate data. Given the
correct hyperparameters, we used mean-field and SSVI-A\footnote{We did
  not compare with SSVI, since its performance on the LDA experiment
  was so similar to that of SSVI-A.} (using the full dataset as a
``minibatch'') to approximate the posterior $p(z, \pi, \phi|y)$. We
also applied collapsed Gibbs sampling (CGS) \citep{Neal:2000}, which
yields samples from the posterior that are asymptotically unbiased.

SSVI-A's performance closely mirrored that of CGS; both methods were
more accurate than mean-field.  Mean-field only discovered 17 of the
56 mixture components; the rest were not significantly associated with
data. By contrast, SSVI-A and CGS used 54 and 55 components
respectively. We also estimated (using Monte Carlo) the KL divergence
between the true data-generating distribution $p(y|\pi,\phi)$ and
$p(y|\hat\pi,\hat\phi)$, where $\hat\pi$ and $\hat\phi$ are the
estimates of the posterior means of $\pi$ and $\phi$ obtained by
mean-field, SSVI-A, and CGS. Mean-field achieved a KL divergence of
5.23, while CGS and SSVI-A achieved much lower KL divergences of 1.9
and 1.94, respectively.




\begin{figure}[t]
  \centerline{\hskip0.2in \includegraphics[width=0.6\textwidth]{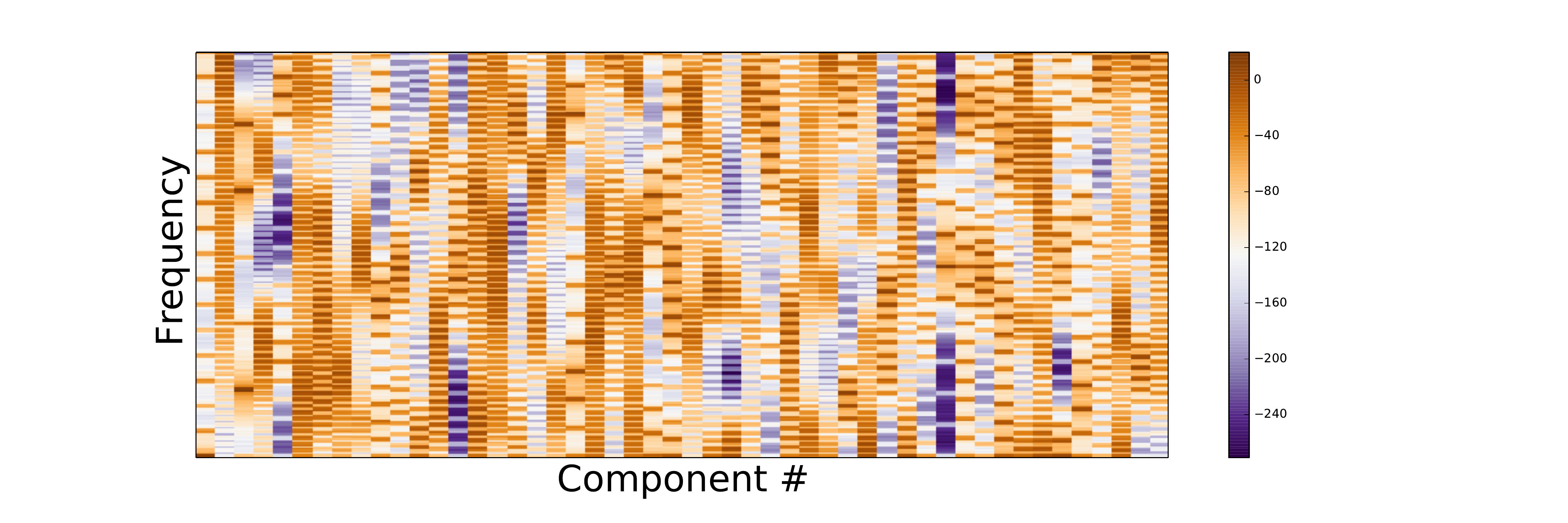}}
  \caption{Dictionary of synthetic spectra used in the GaP-KL-NMF
    experiment. Magnitudes are shown in dB.}
\label{fig:nmfbases}
\end{figure}

\subsection{Bayesian nonparametric nonnegative matrix factorization}
We also evaluated the ability of SSVI-A to determine an appropriate
number of active components in a Bayesian nonparametric model of audio
magnitude spectrograms proposed by \citet{Nakano:2011} as a variant on
the GaP-NMF model of \citet{Hoffman:2010}. The model, which we will
call GaP-KL-NMF, assumes that a quantized magnitude spectrogram matrix
$Y$ is sampled according to the following generative process:
\begin{gather}
W_{f,k}\sim \dgamma(a,a);\quad H_{k,t}\sim\dgamma(b,b) \nonumber
\\ 
\theta_k \sim \dgamma(\alpha/K,\alpha c);
\\\textstyle Y_{f,t} \sim \dpoisson(\sum_k \theta_k W_{f,k} H_{k,t}),
\nonumber
\end{gather}
so that $Y\approx W\mathrm{diag}(\theta)H$.  As $K$ gets large, the
prior on $\theta$ approximates a gamma process, in which most elements
of $\theta$ are expected to be very near 0 but a few may be much
larger than 0. This prior allows the model to determine how many
components it needs to explain the data.

We performed an experiment to compare how well mean-field variational
inference and SSVI-A can accomplish the task of determining how many
latent components actually generated a synthetic spectrogram. We
randomly generated a 257-by-50 dictionary matrix $W$ whose columns
resemble the magnitude spectra of speech sounds (shown in figure
\ref{fig:nmfbases}), sampled a 50-by-1000 activation matrix $H$ of
independent draws from a $\dgamma(0.2, 0.2)$ distribution, and
generated a spectrogram $Y$ such that $Y_{f,t}\sim\dpoisson(\sum_k
W_{f,k} H_{k,t})$. We then used MF and SSVI-A to approximate the
posterior $p(W, H, \theta | Y)$ under the GaP-KL-NMF model with
hyperparameters $a=0.25$, $b=0.2$, $\alpha=1$, $c=1$.

\begin{figure}[t]
  \centerline{\includegraphics[width=0.37\textwidth]{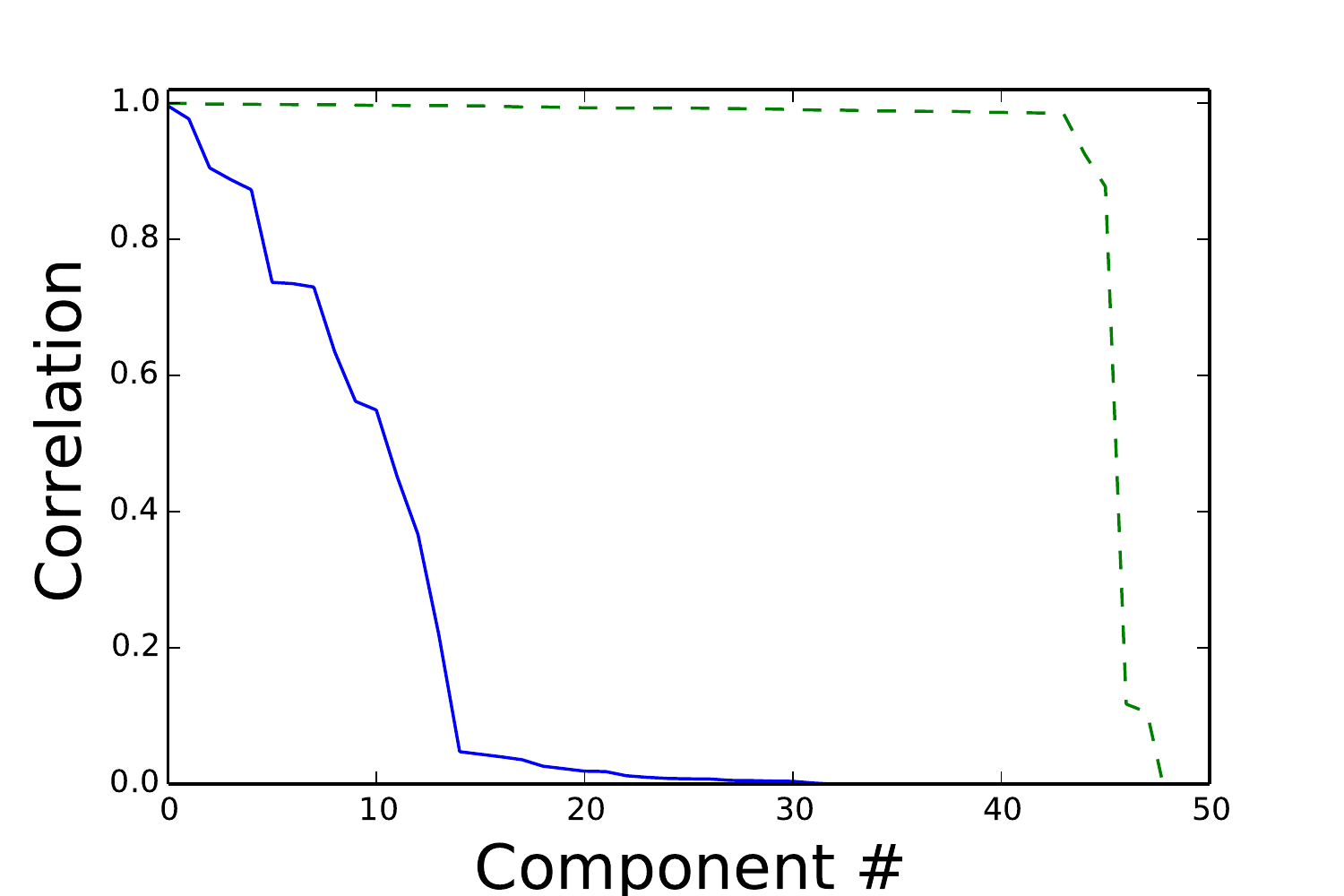}}
  \caption{Sorted correlations between elements of the true dictionary
    of bases used to generate a synthetic spectrogram and the
    dictionaries learned by mean-field (solid blue line) and
    SSVI-A (dashed green line) on GaP-KL-NMF.}
\label{fig:gapnmf_plot}
\end{figure}

Figure \ref{fig:gapnmf_plot} plots the correlation between each column
of the true dictionary $W$ and its closest match among the columns of
the estimated dictionaries $\Eq[W]$ obtained using MF and SSVI-A.
These correlations are sorted in decreasing order. SSVI-A recovers
good approximations of nearly all of the bases used to generate the
data, whereas MF performs quite poorly. This is because MF grossly
underestimates the number of components used to generate the
data---the components with near-zero correlations are not used to
explain the data.  As in section \ref{sec:dpmob}, MF quickly gets
stuck in a bad local optimum from which it cannot recover.





\section{Discussion}
We have presented stochastic structured variational inference (SSVI), an
algorithmic framework that uses stochastic variational inference to
restore the dependencies between global and local unobserved variables
that mean-field variational inference breaks. Experiments suggest that
both SSVI and SSVI-A can fit models in a way that outperforms
previously existing variational inference algorithms.

\appendix

\section{Algorithm derivation}
\label{sec:derivation}

The convergence proofs that our stochastic variational inference
algorithm relies on require that the ELBO be twice continuously
differentiable in $\lambda$ \citep{Hoffman:2013}. We also require that
$q(\beta)$ and $p(y,z,\beta)$ be continuously differentiable in
$\beta$ for any $y$ and $z$. Finally, we require that
$q(z_n|\gamma_n)$ be continuously differentiable in $\gamma_n$, that
$\gamma_n(\beta)$ be continuously differentiable in $\beta$, and
that it be possible to compute an unbiased estimate $\hat\eta_n$
of the expectation $\int_{z_n}q(z_n|\beta)\eta_n(y_n, z_n)dz_n$
for any $\beta$ (for example, using Markov chain Monte Carlo
to sample from $q(z_n|\beta)$).

Our derivation will rely heavily on the concept of \emph{sampling by
  inversion}, a general method for sampling from univariate
distributions. Considering for the moment the case where each
$\beta_k$ is a scalar\footnote{We can also sample by inversion from multivariate distributions such
as the Dirichlet or multivariate normal by generating a set of
independent random variables by inversion and then transforming them,
so that
\vspace{-0.07in}
\begin{equation}
\begin{split}
\label{eq:multivariate}
R(u) = T(\hat R(u)),
\end{split}
\end{equation}
\vskip -0.07in
where $\hat R$ is a vector of $K$ univariate quantile functions and
$T$ is a transformation that introduces dependencies. For example, if
$q$ is a multivariate normal distribution with mean 0 and
covariance $\Sigma$, we could make each $\hat R_k$ the quantile
function of a standard normal and let $T(\hat R)\equiv\sqrt{\Sigma}R$,
so that if $u$ is a vector of uniform random numbers then $T(\hat
R(u))$ would be a draw from $q$. Or, to sample from a Dirichlet we
could use the relation $\hat\beta_k\sim \dgamma(\lambda_k,1)
\Rightarrow
\hat\beta / \sum_l\hat\beta_l\sim\ddirichlet(\lambda)$, set
$\hat R_k(u_k)$ to be the $\dgamma(\lambda_k,1)$ quantile
function, and set $T(\hat R(u))=\hat R(u) / \sum_k\hat
R_k(u)$.}, 
we can obtain a draw of $\beta$
from $q$ by first sampling $K$ uniform random variables
$u_k\sim\duniform([0,1])$ and then passing each through the quantile
function $R$; i.e., $\beta\sim q \Leftrightarrow u_k \sim
\duniform([0, 1]), \beta_k = R_k(u_k)$. We will use the vector-valued
functions $R(u) = (R_1(u_1),\ldots,R_K(u_K))^\top$
$Q(u) = (Q_1(u_1),\ldots,Q_K(u_K))^\top$ to denote the
concatenations of the outputs of the $K$ quantile functions and CDFs. $R$
and $Q$ depend on both $u$ and on the parameters $\lambda$ that control
$q(\beta)$, but we suppress the dependence on $\lambda$ to avoid
clutter. Below, we will need the derivative of $R$ with respect to
$\lambda$; this can be obtained using an identity derived in appendix
\ref{sec:quantilederiv}.

We begin by writing the bound in equation \ref{eq:elbo} as a function
of $\lambda$.  
We define $\cL(\lambda)$ as the ELBO achieved by
setting $q(z_n|\beta)=q(z_n|\gamma_n(\beta))$ for all values of
$\beta$:
\begin{gather}
\label{eq:loflambda}
\textstyle\cL(\lambda)\equiv\textstyle
\int_\beta q(\beta)(\log \frac{p(\beta)}{q(\beta)}
+ \sum_n\cL_{n}(\beta,\gamma_n(\beta))) d\beta;
\\
\cL_{n}(\beta, \gamma) \textstyle\equiv
\int_{z_n}q(z_n|\gamma(\beta))
\log\frac{p(y_n,z_n|\beta)}{q(z_n|\gamma_n(\beta))}dz_n
\end{gather}
where we define $\cL_{n}(\beta,\gamma_n)$ as a shorthand for
the part of the ELBO that depends on $y_n$ and $z_n$.

We will optimize this bound using stochastic optimization. To do so, we need
to consider the derivative of $\cL(\lambda)$ with respect to
$\lambda$. The derivative of the first term in the expectation
simplifies:
\begin{equation}
\begin{split}
&\textstyle\nabla_\lambda\int_\beta q(\beta)\log\frac{p(\beta)}{q(\beta)}d\beta
\\&\textstyle= \nabla_\lambda
((\eta - \lambda)^\top\nabla_\lambda A(\lambda)
- A(\eta) + A(\lambda))
\\&\textstyle= \nabla^2_\lambda A(\lambda)
(\eta - \lambda).
\end{split}
\end{equation}
This is a consequence of the exponential-family identity
$\Eq[t(\beta)]=\nabla_\lambda A(\lambda)$. The derivatives of the
remaining terms do not simplify so easily.  It will be easier to
simplify them by rewriting the expectation of
$\cL_{n}$ in terms of the quantile function $R(u)$
we defined in section \ref{sec:requirements}:
\begin{equation}
\label{eq:lquantile}
\textstyle
\int_\beta q(\beta) \cL_{ n}(\beta,\gamma_n(\beta))d\beta
 = 
\int_u \cL_{n}(R(u),\gamma_n(R(u)))du.
\end{equation}
Now, taking the derivative of $\cL_{ n}$ with respect to
$\lambda$ using the chain rule yields
\begin{align}
\label{eq:partiald}
&\textstyle
\nabla_\lambda \cL_n(R(u),\gamma_n(R(u)))
\\&=
\textstyle(\nabla_\lambda R(u))\Big(\nabla_\beta \cL_n(R(u), \gamma_n(R(u)))
\nonumber
\\&\qquad\qquad\textstyle+ (\nabla_\beta \gamma_n(R(u)))\nabla_{\gamma_n} \cL_n\big(R(u), \gamma_n(R(u))\big)\Big).
\nonumber
\end{align}
The second term vanishes because 
$\nabla_{\gamma_n}\cL_n(\beta, \gamma_n(\beta))=0$ by the
definition in equation \ref{eq:gammastar}, and so the derivative
of equation \ref{eq:lquantile} simplifies to
\begin{align}
\label{eq:simplifiedderivative}
&\textstyle
\nabla_\lambda
\int_u \cL_{n}(R(u),\gamma_n(R(u)))du 
\\&=\textstyle
\int_u (\nabla_\lambda R(u)) \nabla_\beta \cL_n(R(u), \gamma_n(R(u)))du
\nonumber
\\&=\textstyle
\int_{u,z_n}q(z_n|\gamma_n(R(u)))
(\nabla_\lambda R(u))
(\nabla_\beta t(R(u)))^\top
\nonumber
\\&\qquad\qquad\qquad\qquad\qquad\qquad\qquad\qquad\eta_n(y_n, z_n)
dz_n du
\nonumber
\end{align}
Now, if
we sample the global parameters $\beta\sim q_\beta$ and compute an
unbiased estimate $\hat\eta_n$ of the expectation
$\int_{z_n}q(z_n|\gamma_n(\beta))\eta_n(y_n,z_n)dz_n$,
then we can compute a random vector $g$ whose expectation
is the true gradient:
\begin{align}
\label{eq:sg}
 g &\equiv \textstyle
(\nabla^2_\lambda A(\lambda))(\eta - \lambda)
\\&\quad\textstyle+ (\nabla_\lambda R(Q(\beta)))(\nabla_\beta t(\beta))^\top
\sum_n \hat\eta_n;
\quad \E[ g] = \nabla_\lambda\cL.
\nonumber
\end{align}
A stochastic natural gradient can be obtained by preconditioning this
stochastic gradient with the inverse of the Fisher information matrix of
$q(\beta)$:
\begin{equation}
\begin{split}
\label{eq:sng}
 g^{\mathrm{nat}} &\equiv \textstyle
-\lambda + \eta
+ V(\beta, \lambda)\sum_n \hat\eta_n,
\end{split}
\end{equation}
where $V(\beta,\lambda)$ is defined as in equation \ref{eq:vdef} as
\begin{equation}
\begin{split}
V(\beta, \lambda)\equiv 
(\nabla^2_\lambda A(\lambda))^{-1}
(\nabla_\lambda R(Q(\beta)))(\nabla_\beta t(\beta))^\top
\end{split}
\end{equation}
It is easy to verify that $\E[ g^{\mathrm{nat}}] = \textstyle
(\nabla^2_\lambda A(\lambda))^{-1}\nabla_\lambda\cL$. Because the
Fisher information matrix of $q(\beta)$ is the second derivative of
$A$ with respect to $\lambda$, $\E[ g^{\mathrm{nat}}]$ is the natural
gradient of $\cL$ with respect to $\lambda$ \citep{Sato:2001}. We can
therefore optimize $\cL$ with respect to $\lambda$ by repeatedly
sampling values of $ g^{\mathrm{nat}}$ and plugging them into a
standard Robbins-Monro stochastic approximation algorithm
\citep{Hoffman:2013}. Such an approach is summarized in algorithm
\ref{alg:structured} (SSVI). 

\section{Derivatives of Quantile Functions}
\label{sec:quantilederiv}
One definition of the quantile function is as the inverse of the
cumulative distribution function (CDF) $Q_k(\beta_k,
\lambda)=q(\beta_k < \beta_k)$. Writing down the definition of
an inverse function and Differentiating both sides of this definition
shows that
\begin{equation}
\begin{split}
\label{eq:dR}
\textstyle
\partiald{R_k}{u_k}|_{Q_k(\beta_k, \lambda),\lambda}
\partiald{Q_k}{\lambda}&\textstyle|_{\beta_k,\lambda} + 
\partiald{R_k}{\lambda}|_{Q_k(\beta_k, \lambda),\lambda} \textstyle= 0
\\\textstyle \partiald{R_k}{\lambda}|_{Q_k(\beta_k, \lambda),\lambda}
&= \textstyle
-(\partiald{Q_k}{\beta_k}|_{\beta_k,\lambda})^{-1}
\partiald{Q_k}{\lambda}|_{\beta_k,\lambda}
\\\textstyle \partiald{R_k}{\lambda}|_{Q_k(\beta_k, \lambda),\lambda}
&= \textstyle -q(\beta_k)^{-1}\partiald{Q_k}{\lambda}|_{\beta_k,\lambda},
\end{split}
\end{equation}
where we use the identities that the derivative of a function's
inverse is one over the derivative of that function and that the
derivative of a CDF with respect to the random variable is the
corresponding probability distribution function (PDF). The derivative
of $Q_k$ with respect to $\lambda_k$ can be obtained numerically using
finite differences or automatic differentiation. (The same is true of
$R_k$, but CDFs are often much cheaper to compute than quantile
functions.) For multivariate distributions defined as in equation
7 (main text) we can compute $\partiald{R}{\lambda}$ as
\begin{equation}
\begin{split}
\textstyle
\partiald{R}{\lambda}|_{u,\lambda} &= \textstyle
\partiald{T}{\hat R}|_{\hat R(u,\lambda)}
\partiald{\hat R}{\lambda}|_{u, \lambda}
\\ \textstyle\partiald{\hat R_k}{\lambda} &=  \textstyle
-\hat q_k(\hat R(u_k,\lambda))^{-1}
\partiald{\hat Q_k}{\lambda},
\end{split}
\end{equation}
where $\hat q_k$ is the PDF of the $k$th random variable obtained via
the $k$th univariate quantile function $\hat R_k$ and $\hat Q_k$ is
the CDF that is the inverse of $\hat R_k$.

\section{SSVI for Latent Dirichlet Allocation}
In this section we demonstrate how to use SSVI to do approximate
posterior inference on the popular topic model latent Dirichlet
allocation (LDA) \citep{Blei:2003b}. LDA is a generative model of text
that assumes that the words in a corpus of documents are generated
according to the process
\begin{equation}
\begin{split}
\beta_k \sim \ddirichlet(\eta,\ldots,\eta);
\quad &\theta_d \sim \ddirichlet(\alpha,\ldots,\alpha);
\\  z_{d,n} \sim \dmultinomial( \theta_d);\quad & w_{d,n}
\sim \dmultinomial(\beta_{z_{d,n}}),
\end{split}
\end{equation}
where $ w_{n,m}\in\{1,\ldots,V\}$ is the index into the
vocabulary of the $m$th word in the $n$th document, $
z_{n,m}\in\{1,\ldots,K\}$ indicates which topic is responsible for
$ w_{n,m}$, $\theta_{n,k}$ is the prior probability of a
word in document $n$ coming from topic $k$, and $\beta_{k,v}$ is
the probability of drawing the word index $v$ from topic $k$. For
simplicity we use symmetric Dirichlet priors.

LDA fits into the SSVI framework; the
random variables $\beta$, $\theta$, $ z$, and
$ w$ can be broken into global variables ($\beta$) and $N$
sets of local variables ($\theta_n$, $ z_n$, and $
w_n$) that are conditionally independent given the global variables,
and the posterior over $\beta$ given $ w$, $ z$, and
$\theta$ is in the same tractable exponential family as the
prior (i.e., a Dirichlet):
\begin{equation}
\begin{split}
p(\beta|w,z,\theta) &= \textstyle\prod_k \ddirichlet(\beta_k; \eta+c_k)
\\
c_{k,v} &\equiv \textstyle\sum_{n,m} \I[w_{n,m}=v]\I[z_{n,m}=k],
\end{split}
\end{equation}
where $c_{k,v}$ counts the number of times that the word $v$ is
associated with topic $k$. Our goal will be to approximate the
marginal posterior $p(\beta|w)\propto\int_{\theta, z} p(w, z, \theta,
\beta)dzd\theta$ with a product of Dirichlet distributions
$q(\beta)=\prod_k \ddirichlet(\beta_k; \lambda_k)$. Algorithm
1 requires that we be able to sample from
$q_{\beta}$ by inversion, but the Dirichlet distribution lacks a
well-defined quantile function. However, a Dirichlet random variable
can be constructed from a set of independent gamma random variables:
\begin{gather}
\textstyle \beta'_{k,v}\sim\dgamma(\lambda_{k,v},1);\quad
\beta_{k,v} = \frac{\beta'_{k,v}}{\sum_i\beta'_{k,i}} \nonumber
\\
\textstyle\Rightarrow  \beta_k \sim \ddirichlet(\lambda_{k,1},\ldots,\lambda_{k,v}).
\end{gather}
So we could sample from $q_{\beta}$ by sampling $KV$ independent
uniform random variables $u_{k,v}$, passing each through the gamma
quantile function $\hat R$ to get $\beta_{k,v}'\equiv \hat R(u_{k,v},
\lambda_{k,v}, 1)$, and letting
$\beta_{k,v}=\beta'_{k,v}/\sum_i\beta'_{k,i}$ so that we have
$R(u_{k,v}, \lambda_{k,v}) \equiv \hat R(u_{k,v},\lambda_{k,v},1)
/ \sum_{i} \hat R(u_{k,i},\lambda_{k,i},1).$

\begin{figure*}[t]
  \centerline{\includegraphics[width=0.95\textwidth]{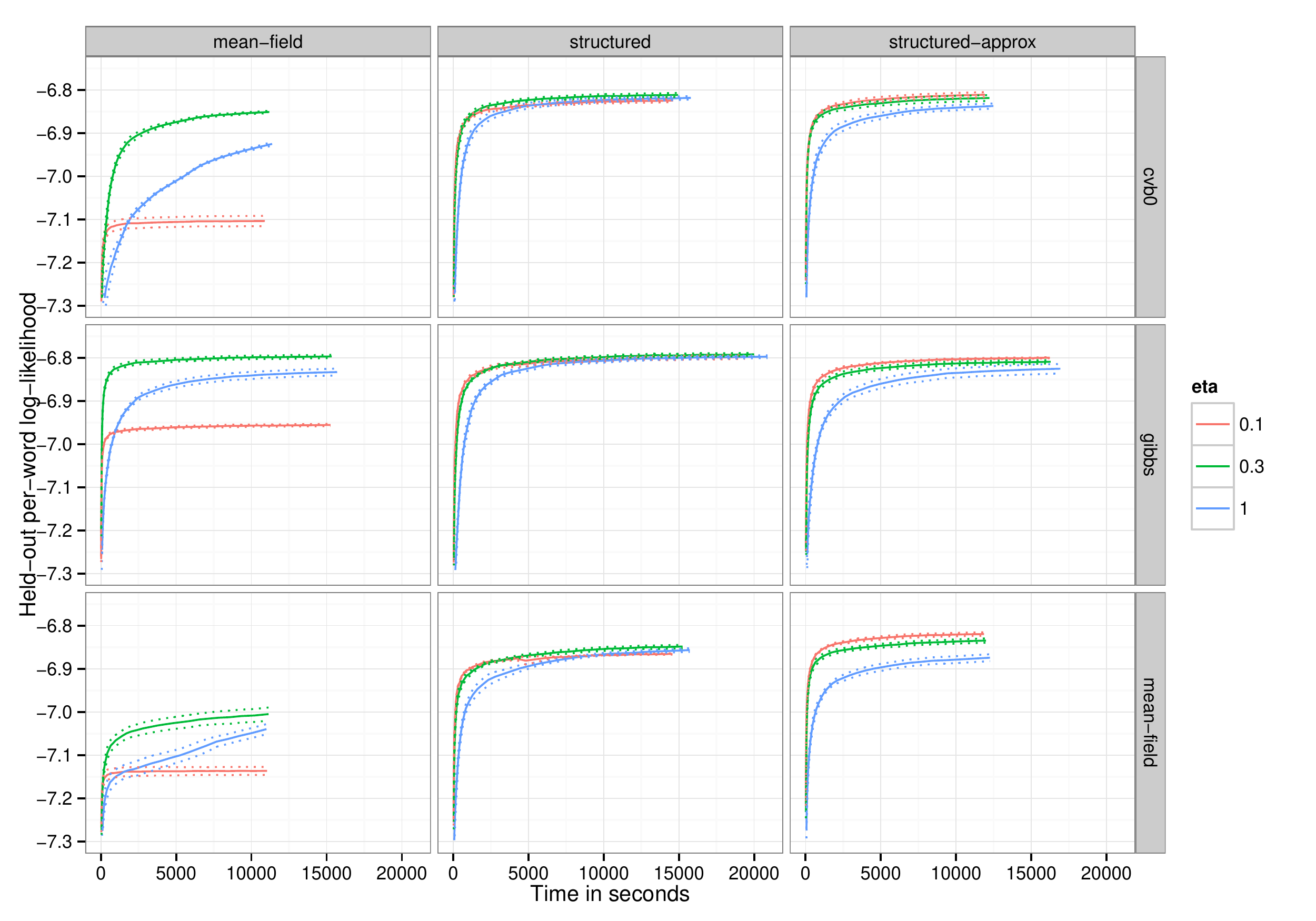}}
  \caption{Predictive accuracy for various algorithms as a function of
    wallclock time when fitting LDA to 3.8 million Wikipedia
    articles. Each algorithm ran two sweeps over the dataset. Solid
    lines show average performance across five runs, dotted lines are
    drawn one standard deviation above and below the mean. The
    algorithms used to update the global parameters and local
    conditional distributions vary horizontally and vertically,
    respectively.}
\label{fig:ldaplotfull}
\end{figure*}

To compute the update for $\lambda$ in algorithm 1
we need to know $(\frac{\partial^2
  A}{\partial\lambda\partial\lambda^\top})^{-1}
(\partiald{t}{\beta}|_{\beta^{(t)}}
\partiald{R}{\lambda}|_{u^{(t)},\lambda^{(t)}})^\top \eta_{
  n}(w_{n^{(t)}}, z^{(t)})$. Since each $\beta_k$ is independent of
all of the other topic vectors under $q$, we need only consider a
single $\beta_k$ at a time. The sufficient statistic vector for the
Dirichlet distribution is $t(\beta_k) = \log \beta_k$, so we have
\begin{multline}
\textstyle
\partiald{t}{\beta_k}|_{\beta_k^{(t)}}
\partiald{R}{\lambda_k}|_{u_k^{(t)},\lambda_k^{(t)}}
\\\textstyle= \mathrm{diag}(\beta_k)^{-1}(\sum_v \beta'_{k,v})^{-1}
(I - \beta 1^\top)\partiald{\hat R}{\lambda_k}|_{u_k, \lambda_k}.
\end{multline}
$\partiald{\hat R}{\lambda_k}$ can be evaluated using equation
\ref{eq:dR}.
$\eta_{ n}(w, z)$ is simply a matrix counting how many times
each unique word is associated with each topic:
$\eta_{ n}(w, z)_{k,v} = \sum_m \mathbb{I}[w_m=v]\mathbb{I}[z_m=k].$
Finally, the log-normalizer for the Dirichlet is
$A(\lambda_k) = -\log\Gamma(\sum_v\lambda_{k,v}) +
\sum_v\log\Gamma(\lambda_{k,v}),$
and the Fisher matrix $\frac{\partial^2 A}{\partial \lambda_k \partial
  \lambda_k^\top}$ is a diagonal matrix plus a rank-one matrix:
$\textstyle\frac{\partial^2 A}{\partial \lambda_k \partial
  \lambda_k^\top} = \mathrm{diag}(\Psi'(\lambda_k)) -
\Psi'(\sum_v \lambda_{k,v}) 1 1^\top,$
where $1$ is a column vector of ones and $\Psi'$ is the second
derivative of the logarithm of the gamma function. The product of the
inverse of the Fisher matrix and a vector can therefore be computed in
$O(V)$ time using the matrix inversion lemma \citep{Minka:2000}.

We now have everything we need to apply algorithm 1 to LDA.

\section{Full Matrix of LDA Results}
\label{sec:ldamatrix}

We tested various combinations of E-steps and M-steps for latent
Dirichlet allocation with 100 topics on the 3,800,000-document
Wikipedia dataset from \citep{Hoffman:2013}. To update the global
variational distributions, we used traditional mean-field updates,
SSVI updates, and SSVI-A updates. For the local variational
distributions, we used the traditional mean-field approximation
\citep{Blei:2003b}, the CVB0 algorithm of \citet{Asuncion:2009}, and
Gibbs sampling as in \citep{Mimno:2012}. We also experimented with
various settings of the hyperparameters $\alpha$ and $\eta$, which
mean-field variational inference for LDA is known to be quite
sensitive to \citep{Asuncion:2009}. For all algorithms we used
mini-batches of 1000 documents and a step size schedule
$\rho^{(t)}=t^{-0.75}$.

Figure \ref{fig:ldaplotfull} summarizes the results for $\alpha=0.1$,
which yielded the best results for all variational algorithms. Using
traditional mean-field inference (bottom row) to approximate
$p(z_n|y_n, \beta)$ degrades performance, but the CVB0 approximation
(top row) works almost as well as Gibbs sampling (middle row) for the
two SSVI algorithms. CVB0 is outperformed by Gibbs when using the
mean-field M-step. The two SSVI algorithms perform comparably well,
but the mean-field M-step (left column) is very sensitive to
hyperparameter selection compared to SSVI and SSVI-A.

\bibliography{bib}
\bibliographystyle{apalike}

\end{document}